\pdfoutput=1

\documentclass[11pt]{article}

\usepackage[preprint]{coling}

\usepackage{times}
\usepackage{latexsym}

\usepackage[T1]{fontenc}

\usepackage[utf8]{inputenc}

\usepackage{microtype}

\usepackage{inconsolata}

\usepackage{graphicx}

\usepackage{subcaption}
\usepackage{amsmath}
\usepackage{amssymb}
\usepackage{algorithm}
\usepackage{algpseudocode}
\usepackage{multirow}

\title{A Robust Autoencoder Ensemble-Based Approach for Anomaly Detection in Text}

\author{Jeremie Pantin \and Christophe Marsala \\
  Sorbonne Universit\'e - LIP6 \\
  Paris - France \\
  \texttt{Firstname.Lastname@lip6.fr} \\}

\begin{document}
\maketitle
\begin{abstract}
Anomaly detection (AD) is a fast growing and popular domain among established applications like vision and time series.
We observe a rich literature for these applications, but anomaly detection in text is only starting to blossom.
Recently, self-supervised methods with self-attention mechanism have been the most popular choice.
While recent works have proposed a working ground for building and benchmarking state of the art approaches, we propose two principal contributions in this paper: contextual anomaly contamination and a novel ensemble-based approach.
Our method, Textual Anomaly Contamination (TAC), allows to contaminate inlier classes with either independent or contextual anomalies.
In the literature, it appears that this distinction is not performed.
For finding contextual anomalies, we propose RoSAE, a Robust Subspace Local Recovery Autoencoder Ensemble.
All autoencoders of the ensemble present a different latent representation through local manifold learning.
Benchmark shows that our approach outperforms recent works on both independent and contextual anomalies, while being more robust.
We also provide 8 dataset comparison instead of only relying to Reuters and 20 Newsgroups corpora.
\end{abstract}

\section{Introduction}
Anomaly Detection (AD) \cite{chandola_anomaly_2009,ruff_unifying_2021} and Outlier Detection (OD) \cite{hodge_survey_2004,zhang_advancements_2013,aggarwal_outlier_2017} are tasks aiming to estimate whether an observation  is normal or not.
These tasks have been recently applied on text for highlighting particular or abnormal texts in a corpus.
In this context, it is still a blossoming and promising field.
Nowadays, most popular approaches are performing One-Class Classification (OCC) for discerning outliers (anomalies) from inliers \cite{kannan_outlier_2017,lai_robust_2020,fouche2020mining}, and self-supervised approaches are often preferred for text \cite{ruff-etal-2019-self,manolache2021date,das2023few}.
Self-supervised approaches are not limited to text, and applications with large language model for vision exists \cite{wang2019effective,zhou2023anomalyclip}.

\begin{figure}[t!]
	\centering
	\includegraphics[width=.45\textwidth]{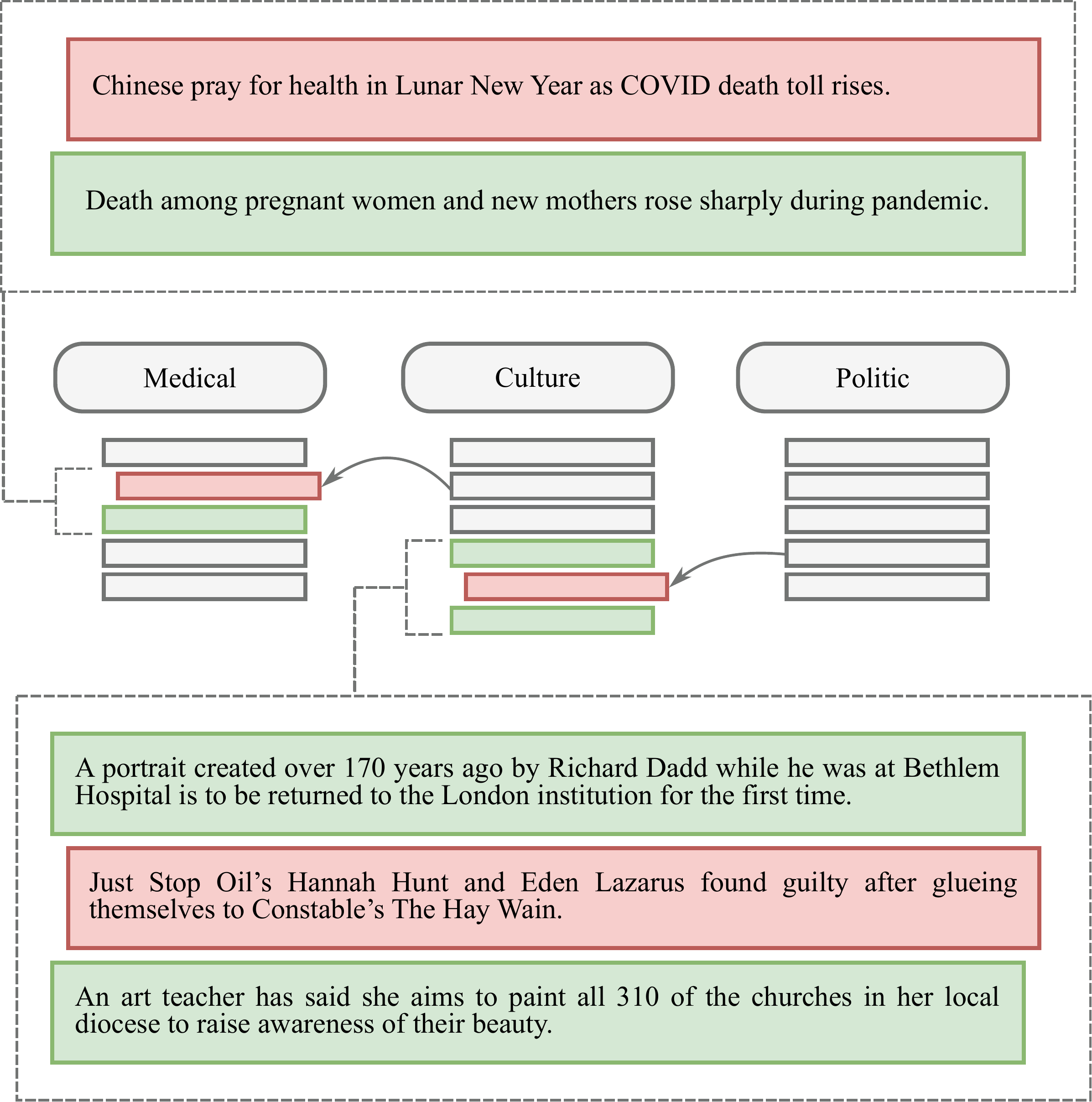}
	\caption{Presentation of the studied problem with three documents topics: medical, culture and politic. Under each topic we represent a textual document with colored rectangles. Gray and green are inliers and red ones are anomalies. The detailed documents are the abstract of the news articles taken from sources like Reuters, New York times, BBC, ... The first scenario is the apparition of a culture-related document in a medical feed, and the second scenario is a political document in the culture feed.}
    \label{fig:text_outlier_example}
\end{figure}

Ensemble-based approaches are not present in the literature, despite their great success on various kind of data \cite{chen2017outlier,briskilal2022ensemble}.
Whereas those approaches benefits of decision robustness and reduced variance.
Most of recent approaches are using Multi-Head Self-Attention \cite{ruff-etal-2019-self,manolache2021date,das2023few}, learning a representation from pre-trained language models as BERT-based models \cite{devlin_bert_2019,liu2019roberta,reimers-2019-sentence-bert}.
An underlying problem of AD lies in the identification of different kinds of anomalies.
Recent approaches does not usually consider these differences when building a detection model.
While independent, contextual and collective anomalies have been identified for other kind of data \cite{chandola_anomaly_2009,aggarwal_categorical_2017}, it is different for text.
Usually text anomalies are only considered as documents from one topic appearing in the wrong topic subset (Figure~\ref{fig:text_outlier_example}).
However anomalies can have multiple forms, and independent anomalies are easier to find than contextual anomalies.

In this paper we introduce a novel ensemble-based approach that performs contextual anomaly detection.
In comparison to state-of-the-art approaches, we focus first on the identification of two kinds of anomaly, independent and contextual, and introduce the Textual Anomaly Contamination (TAC), a novel experimental setup.
Also, our main contribution lies in the introduction of a novel approach for detecting anomalies which is based on a robust autoencoder ensemble.
Our autoencoder, Robust Subspace Local Recovery AutoEncoder (RLAE), faces contaminated corpora (trained on both inliers and anomalies) and discriminates inliers from anomalies under the assumption that one distribution is highly contaminated.
Thus, our Robust Subspace Local Recovery Autoencoder Ensemble (RoSAE) represents inliers in a low-dimensional subspace and proposes different manifold projections for one observation.

\section{Related works} \label{sec:related}
Recent advances in word embedding with language models like BERT \cite{devlin_bert_2019} or RoBERTa \cite{liu2019roberta}, and large language models such as Llama \cite{dubey2024llama} and OPT \cite{zhang2022opt} have shown promising characteristics for anomaly detection.
However, only few methods of the literature propose their use \cite{ruff-etal-2019-self,manolache2021date,das2023few}.
Other methods like One-Class Support Vector Machine (OCSVM) \cite{scholkpof-etal-2001-estimating} and Textual Outlier using Nonnegative Matrix Factorization (TONMF) \cite{kannan_outlier_2017} rely on a TF-IDF representation.
Surprisingly, recent methods are not using outlier ensemble methods \cite{aggarwal_theoretical_2015,zhao2019lscp,zimek2014ensembles} for performing anomaly/outlier detection with text data.

AutoEncoders (AE) have been widely used for anomaly/outlier detection with high-dimensional data \cite{chen2017outlier,kieu2019outlier} and are also successful with other kind of data \cite{an2015variational,chen2018autoencoder,lai_robust_2020,zhou2017anomaly,beggel2020robust}.
The risk of using autoencoders with language models leads to the apparition of degenerate solution in the learning step.
Robust properties are needed in such scenario for handling the manifold collapse phenomenon \cite{wang-wang-2019-riemannian}.
Such issue can be mitigated using robust projections in the learned space of the autoencoder (eigenfunction decomposition \cite{bengio2004learning}, ...).
The Robust Subspace Recovery \cite{lerman2018overview,rahmani2017randomized} is a robust manifold learning technique that map inlier distributions in a subspace where anomalies/outliers are at the edge.
Based on such an approach, an autoencoder \cite{lai_robust_2020} encodes observation from the input space in a robust subspace.
One problem of such autoencoder is the difficulty to address scenarios with features that present similar range values (problem of entanglement) \cite{wang2022disentangled}.
Entanglement is especially more difficult with contextual anomalies where documents shares similar semantic properties.
Such problem can be tackled using a local projection from the nearest neighbours in the latent representation space.
Several manifold learning approaches like locally linear embedding (LLE) \cite{roweis2000nonlinear}, neighbourhood components analysis (NCA) \cite{goldberger2004neighbourhood} or soft nearest neighbours (SNN) \cite{frosst2019analyzing} can help to mitigate it.

\begin{figure*}[t!]
    \centering
    \includegraphics[width=.85\textwidth]{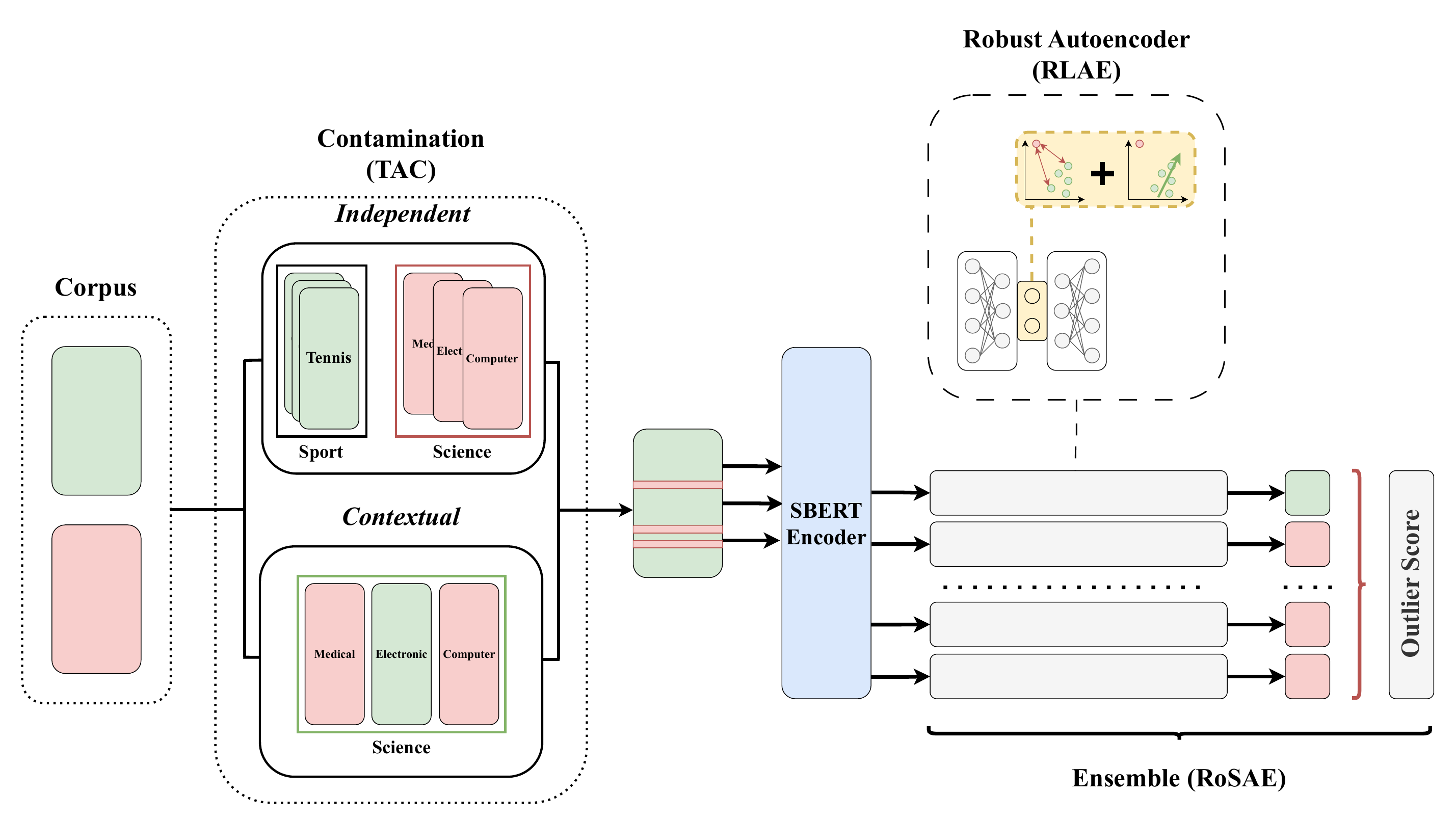}
    \caption{Overview of our global system, including contamination setup and our proposed approach RoSAE.}
    \label{fig:contamination}
\end{figure*}

A large number of contributions have settled anomaly/outlier taxonomies \cite{zhang_advancements_2013,aggarwal_categorical_2017,ruff-etal-2019-self} and several types of anomalies have been proposed in the literature: Independent anomaly, Contextual anomaly and Collective anomaly.
Consequently, various types of anomalies frequently coexist within one corpus.
A topic is defined as the main subject of a document \cite{bejan-etal-2023-ad}.
Depending on the document, there may be multiple subtopics within a broader category (e.g., a sports topic that encompasses football and tennis).
Thus, accounting for this hierarchical structure, a form of contextual anomaly arises \cite{fouche2020mining}.

Three different approaches have been used for performing corpus contamination.
The first is a manually picked contamination \cite{manevitz_one-class_2001,kannan_outlier_2017}, using text classification corpora like 20 Newsgroups, where topics are unrelated.
A more practical contamination process is performed by recent works like \citet{lai_robust_2020,fouche2020mining}, in which the inlier topic is selected and all other topics are considered anomalies.
In this scenario, the hierarchical structure of topics is not handled, and related topics are mixed with each others. 
Finally, more recently, corpora have been prepared with careful consideration of topic hierarchy \cite{ruff-etal-2019-self,manolache2021date,das2023few}.
We also note that there exists different kind of training procedure.
In \citet{ruff-etal-2019-self,manolache2021date} the proposed model is trained on inlier class, and then tested on contaminated split.
For \citet{lai_robust_2020,fouche2020mining,das2023few}, the training and testing step is performed on contaminated split.

\section{RoSAE: Robust Subspace Local Recovery Autoencoder Ensemble} \label{sec:erla}
While robust subspace recovery autoencoders have successfully tackled anomaly detection in text corpora, they lack locality and geometry awareness for mitigating disentanglement and manifold collapse in transformer-based language models.
Thus, we introduce Robust Subspace Local Recovery AutoEncoder (RLAE) which integrates locality in the latent representation through a local neighbouring method before presenting our ensemble-based approach RoSAE.

\subsection{Randomly Connected One-Class Autoencoder}
\label{subsec:randomAE}
Let $\mathrm{X}$ be a dataset of $\mathrm{N}$ instances such as $\mathrm{X} = \{\mathrm{x_1}, ..., \mathrm{x_N}\}$.
Each instance has $\mathrm{D}$ dimensions which correspond to its attributes: $\mathrm{x_i} = \{x_1, ..., x_D\}$.
An Autoencoder is a neural networks in which the encoder $\mathcal{E}$ maps an instance $\mathrm{x_i}$ in a latent representation noted $\mathrm{z}_i = \mathcal{E}(\mathrm{x}_i) \in \mathbb{R}^e$ of dimension $e$.
The Robust Subspace Recovery (RSR) layer is a linear transformation $\mathbf{A}\in\mathbb{R}^{d \times e}$ that reduces the dimension to $d$.
We denote $\mathrm{\hat{z}_i}$ the representation of $\mathrm{z_i}$ through the RSR layer, such as $\mathrm{\hat{z}_i} = \mathbf{A}\mathrm{z_i} \in \mathbb{R}^d$.
The decoder $\mathcal{D}$ maps $\mathrm{\hat{z}_i}$ to $\mathrm{\hat{x}_i}$ in the original space $D$.
The matrix $\mathbf{A}$ and the parameters of $\mathcal{E}$ and $\mathcal{D}$ are obtained with the minimization of a loss function.

Similarly to \cite{chen2017outlier} we introduce an autoencoder with random connection such as we increase the variance of our model.
In the autoencoder ensembles, each autoencoder has a random probability of having several of its connections to be pruned.
Thus, we set the probability of disconnection with a random rate in $[0.2,0.5]$.

\subsection{Robust Subspace Recovery Layer}
\label{subsec:rsr}
The RSR layer maps the inliers around their original locations, while outliers are projected far from their original locations \cite{lerman2018overview,lai_robust_2020}.
The loss function minimises the sum of the autoencoder loss function noted $L_{AE}$ with the RSR loss function noted $L_{RSR}$.
\begin{equation}
    L^p_{AE}(\mathcal{E},\mathbf{A},\mathcal{D}) = \sum^{N}_{i=1}{||\mathrm{x_i} - \mathrm{\hat{x}_i}||^p_2}
\end{equation}
which is the $l_{2,p}-norm$ based loss function for $p > 0$.

For performing the subspace recovery, we denote two terms that have different roles in the minimisation process.
Two terms are presented, with the first term enforcing the RSR layer to be robust (PCA estimation), and the second enforcing the projection to be orthogonal:
\begin{align}\nonumber
L^q_{RSR}(\mathbf{A}) = & \lambda_1 \sum^{N}_{i=1}{||\mathrm{z_i} - \mathbf{A}^{\mathrm{T}}\mathrm{\hat{z}_i}||^q_2} \\
& + \lambda_2 \sum^{N}_{i=1}{||\mathbf{AA}^{\top} - \mathbf{I}_d||^q_f}
\label{eq:sec4_rsr_loss}
\end{align}
with $\mathbf{A}^{\top}$ the transpose of $\mathbf{A}$, $\mathbf{I}_d$ the $d \times d$ matrix and $||\cdot||_f$ the Frobenius norm. $\lambda_1$ and $\lambda_2$ are hyperparameters and $q=1$ is corresponding to the optimal $l_{p,q}$ norm \cite{maunu2019well}.
If we simplify Equation~\ref{eq:sec4_rsr_loss} we have:
\begin{align}
	L_{RSRAE}(\mathcal{E}, \mathbf{A}, \mathcal{D}) = & \lambda_1 L^1_{AE}(\mathcal{E}, \mathbf{A}, \mathcal{D}) \nonumber\\
 &  + \lambda_2 L^1_{RSR}(\mathbf{A})
\end{align}

\subsection{Robust Local Embedding}
\label{subsec:le}
Locally Linear Embedding (LLE) \cite{roweis2000nonlinear,chen2011locally} proposes the assumption that a local geometry can be linearly represented with surrounding observations.
The Local Neighbouring Embedding (LNE) term in the loss function encourages the autoencoder to learn representations that preserve the relationships between data points and their local neighbours.
Based on Equation~\ref{eq:sec4_rsr_loss}, the reconstruction loss function enforces robustness with $L^1_{AE}$ and the orthogonality with $L^1_{RSR}$.
Because the learned representation of the encoder is compressed in a $e$ dimension space, the locality of the subspace is not handled.

Based on LNE, we introduce a third term.
Given a set of data points $\{\mathrm{x_i}\}_{i=1}^N$ in the input space, the goal of LNE is to find a lower-dimensional representation $\{\mathrm{z_i}\}_{i=1}^N$ in the output space (the subspace learned by the autoencoder) such that the local relationships between data points are preserved.
We note:
\begin{equation}
    L_{LNE}(\mathbf{A}) = \sum_{i=1}^N{\left\lVert \mathrm{x_i} - \sum_{j\in \mathcal{N}_i} w_{ij} \mathrm{x_j} \right\rVert_2^2}
\end{equation}
where $\mathcal{N}_i$ represents the set of indices of the k-nearest neighbours of $\mathrm{x_i}$ (excluding $\mathrm{x_i}$ itself) and $w_{ij}$ are the weights assigned to the neighbouring observations $\mathrm{x_j}$ in the reconstruction of $\mathrm{x_i}$.
The weights $w_{ij}$ can be computed using the least squares method to minimize the reconstruction error: $\min_{\mathbf{w_i}} \left\lVert \mathrm{x_i} - \sum_{j\in \mathcal{N}_i} w_{ij} \mathrm{x_j} \right\rVert_2^2$ subject to the constraint $\sum_{j\in \mathcal{N}_i} w_{ij} = 1$.

The LNE term encourages the autoencoder to find a representation for each data point as a combination of its k-nearest neighbours in the latent space.
With minimisation of the LNE term in the loss function, the autoencoder learns to preserve the local linear relationships, which ultimately helps to project the Euclidean distance with its neighbours.
The reconstruction errors of LNE is measured by the cost function:
\begin{equation}
    L_{LNE}(\mathbf{A}) = \sum^N_{i=1}{\sum_{j\in \mathcal{N}_i} w_{ij} \lVert \textbf{A}\mathrm{x_i} - \mathbf{A}\mathrm{x_j} \rVert_2^2}
\end{equation}
The weight $w_j$ assigned to the neighbour $\mathrm{x_{ij}}$ in the local  reconstruction of $\mathrm{x_i}$ is determined based on its corresponding distances.
The inclusion of the LNE term in the loss function encourages the autoencoder to preserve the local geometric structure of the data in the latent subspace.

Finally, the Robust Subspace Local Recovery AutoEncoder (RLAE) loss function is measured as follows:
\begin{align}
\nonumber
L_{RLAE}(\mathcal{E}, \mathbf{A}, \mathcal{D}) & = L_{RSRAE}(\mathcal{E}, \mathbf{A}, \mathcal{D}) \\
 & \mbox{ } + \lambda_3L_{LNE}(\mathbf{A})
\label{eq:sec4_erla}
\end{align}
The parameter $\lambda_3$ controls the influence of the LNE term on the overall loss.
Because it controls the influence of locality of the manifold, the term is preferred to be low for avoiding degenerate results.

\subsection{Ensemble Learning}
The main idea behind ensemble methods is that a combination of several models, also called \textit{base detectors}, is more robust than usage of a single model.
In RoSAE, we use the reconstruction error of each autoencoders and then we normalise each base detector scores through the standard deviation of one unit.
We then take the median value for each observation.

\subsection{Text Representation}
Recent works have recorded their results on BERT-based language models.
The RoSAE model is not based on the self-attention mechanism, such as for \citet{ruff-etal-2019-self,manolache2021date}, and we use the RoBERTa Sentence Transformer from \citet{reimers-2019-sentence-bert}.

\begin{algorithm}[t]
    \caption{TAC: Textual Anomaly Contamination (Independent)}
    \label{alg:tac}
    \begin{algorithmic}
        \Require Inlier topic $\zeta$, corpus $\mathrm{X}$, split size $l$, contamination rate $\mathrm{\nu}$
        \Ensure $0 < l \leq N $ 
        \State $c \gets l\nu$
        \State $i \gets 0$
        \State Initialize empty matrix $\mathrm{Z}$
        \State $\mathcal{A} \gets \{\mathrm{x}_j \times \mathrm{y}_j \in \mathrm{X} \times \mathrm{Y} | \forall j \in [0,N], \mathrm{y}_j \neq \zeta\}$
        \State $\mathrm{X}_{\zeta} \gets \{\mathrm{X} \backslash \mathcal{A}\}$
        \Comment{Inlier Matrix}
        \While{$|\mathrm{Z}| < c$}\\
        \Comment{Set = instead of $\neq$ for contextual}
        \If{$\mbox{Parent}(\mathrm{y}_i) \neq \mbox{Parent}(\zeta)\mbox{}$}
        \State $\mbox{Append}(\mathrm{x}_i,\mathrm{y}_i)$ to $\mathrm{Z}$
        \EndIf
        \State $i \gets i + 1$
        \EndWhile
        \State Fill $\mathrm{Z}$ with $\mathrm{X}_{\zeta}$ until $|\mathrm{Z}| = l$
        \State \Return $\mbox{Shuffle}(\mathrm{Z})$
    \end{algorithmic}
\end{algorithm}

\section{Textual Anomaly Contamination (TAC)}
Text AD is lacking dedicated dataset, resulting in all state-of-the-art methods preparing benchmark corpora with synthetic contamination.
We propose a contamination algorithm for performing contamination with independent and contextual anomalies (Figure~\ref{fig:contamination}).
To illustrate this, let a legal document mentioning a football player, it would be anomalous if incorrectly appearing in a sports-related corpus.
Independent anomalies represent observations that lack any meaningful relationship with other topics.
Specifically, anomaly topics and inlier topics have different hierarchical parents within the category structure.
Let a corpus of text $\mathrm{X}$ with its labelled topic $\mathcal{Y}$, a labelled document of a corpus $(\mathrm{x},y) \in \mathrm{X}\times\mathcal{Y}$ and $\zeta$ the inlier category
The corresponding inlier subset is noted $\mathrm{X}_\zeta \subseteq \mathrm{X}$.
We define $\mathcal{A}$ the subset of all anomalies such as $\mathcal{A} \subset \mathrm{X}$.
We have:

\begin{equation}
    \mathcal{A} = \mathrm{X} \backslash  \mathrm{X}_\zeta
    \label{eq:generic}
\end{equation}

\noindent
Regarding $\mathcal{A}$, we can make the distinction with two different constraints.
We note $P(y)$ the direct parent of $y$ in a given hierarchy.
First, an observation $\mathrm{x_i}$ is considered to be an anomaly if its parent topic is different of inlier parent topics such as:
\begin{equation}
    \label{eq:point}
    \mathcal{A}_{p}(\zeta) = \{a | \mbox{P}(\zeta) \neq \mbox{P}(\mathrm{y}), (\mathrm{a,y}) \in  \mathcal{A} \times \mathrm{Y}\}
\end{equation}

\noindent
The second constraint corresponds to documents that do not lie in $\mathrm{X}_\zeta$ but share the same parent topic as $\zeta$.
These observations are identified as another kind of anomaly: contextual anomaly.
We write:

\begin{equation}
    \label{eq:contextual}
    \mathcal{A}_{c}(\zeta) = \{a | \mbox{P}(\zeta) = \mbox{P}(\mathrm{y}) , (\mathrm{a,y}) \in  \mathcal{A} \times \mathrm{Y}\}
\end{equation}

The Algorithm~\ref{alg:tac} presents TAC for independent anomalies and considers three hyperparameter: the contamination rate $\nu$, length of the contaminated split $l$ and the inlier topic noted $\zeta$.

\section{Experiments}\label{sec:expe}
In this section we present several experiments on both independent anomalies and contextual anomalies.
We present all corpora and how we perform this study.
Results are commented and an ablation study is proposed.

\subsection{Data}
\label{sec:data}
Recent work \cite{lai_robust_2020,ruff-etal-2019-self,kannan_outlier_2017,mahapatra_contextual_2012} uses classification datasets such as Reuters-21578\footnote{
\url{http://www.daviddlewis.com/resources/testcollections/reuters21578/}
} and 20 Newsgroups\footnote{
\url{http://qwone.com/~jason/20Newsgroups/}
} with a dedicated preparation in order to compare their approaches.
For all corpora we apply data contamination of independent anomalies and contextual anomalies with TAC (Algorithm~\ref{alg:tac}).
To be fair with each method and each dataset, we set the size of the preparation subset to the maximum number of available document, and results are averaged over $10$ of runs.
Data is pre-processed with lowercase and stop words removal.
The train split of each corpus is used for training, and test split for evaluation.
In a first step, we set $\nu = 0.10$.

\paragraph{20 Newsgroups.}
For 20 Newsgroups we separate the subtopics into seven main topics: computer, forsale, motors, politics, religion, science, sports.
\paragraph{Reuters-21578.}
The Reuters-21578 corpus contains documents associated with several topics.
We delete all these documents to keep only those associated with a single topic.
We reorganise the topics in order to obtain a hierarchy, based on the work of \cite{toutanova2001text}.
Thus, four parent themes are created: commodities, finance, metals and energy.
We apply TAC to the eight topics that have the greatest number of training documents.
\paragraph{DBpedia 14.}
For DBpedia 14 \cite{zhang2015character} we create the topic hierarchy based on the provided ontology\footnote{\url{mappings.dbpedia.org/server/ontology/classes/}} and denote six parent topics.
\paragraph{Web Of Science.}
Web of Science \cite{kowsari2017hdltex} is often used as a reference for hierarchical classification and provides three levels of topic hierarchy.
Thus, seven parent topics are present and for child topics that are associated with more than one parent, we only keep the largest set of children.
\paragraph{Spam.}
We use Enron\footnote{\url{https://www.cs.cmu.edu/~enron/}} dataset and SMS Spam \cite{Almeida2011SpamFiltering} dataset.
We prepare a train and test split before contaminating ham in spam, and then spam in ham.
\paragraph{Sentiment.}
IMDB \cite{maas2011learning} and SST2 \cite{socher2013recursive} are used for benchmarking independent contamination with more difficult scenarios.
Preparation of both corpora is similarly performed than for Spam preparation.

\subsection{Preparation}
We use TAC for preparing contextual contamination on each candidate inliers with $\nu = 0.1$.
The maximum amount of available documents is retrieved $|D|$.
All results are comapred against Area Under Curve ROC (AUC) and Average Precision (AP) metrics.
We also keep the number of runs for each corpus and each split contamination to $10$.

\paragraph{Baseline}
We propose to compare our approach against OCSVM\cite{scholkpof-etal-2001-estimating}, RSRAE\cite{lai_robust_2020}, an autoencoder ensemble model RandNet \cite{chen2017outlier}, CVDD \cite{ruff-etal-2019-self} and FATE \cite{das2023few}.
For OCSVM we use the implementation from PyOD\cite{zhao2019pyod} and for RSRAE we use our PyTorch implementation of their code.
All our results approximate theirs.

We also display CVDD results with several adaptations.
In the original paper, CVDD learns on all the training corpus (inlier only) instead of the training split only.
Considering this property, we use their implementation knowing the approach has advantage against other baseline models.
The same comment goes for FATE, which additionally use few-shot learning and benefits from anomaly awareness.

\paragraph{Implementation}
For RoSAE and RandNet we set the number of base predictors to $20$.
We provide the code of our approach\footnote{\url{https://anonymous.4open.science/r/RoSAE-COLING-B12E}} using the PyOD base implementation \cite{zhao2019pyod}.
We propose an implementation based on the \textit{BaseDetector} of PyOD so each of the compared model can be accordingly tested.
We use the pruning methods from PyTorch for creating randomly connected autoencoders.
Additionally, we also set hyperparameters $\lambda_1 = 0.1$, $\lambda_2 = 0.1$ and $\lambda_3 = 0.05$.
For avoiding manifold collapse problem and degenerates solutions, we advise that $\lambda_3 < \lambda_1 $.
We set the latent dimension of the LNE layer to $32$.
On the other hand, we set the epoch number to $50$ and random connection probability between $[0.2,0.5]$. 

\begin{table*}[t]
    \centering
    \begin{tabular}{lcccccccc}
 \hline
 \multicolumn{9}{c}{\textbf{Independent anomalies ($\nu = 0.1$)}}\\
        \hline
        \multicolumn{1}{l}{\textbf{Model}} & \textbf{20New.} & \textbf{Reuters} & \textbf{WOS} & \textbf{DBpedia} & \textbf{Enron} & \textbf{SMS Sp.} & \textbf{IMDB} & \textbf{SST2} \\
        \hline
        OCSVM & 0.884 & 0.851 & 0.876 & 0.962 & 0.687 & 0.683 & 0.539 & 0.575 \\
        RandNet & 0.622 & 0.689 & 0.636 & 0.575 & 0.466 & 0.687 & 0.503 & 0.486 \\
        RSRAE & 0.669 & 0.768 & 0.731 & 0.634 & 0.518 & 0.642 & 0.540 & 0.577 \\
        CVDD & 0.781& 0.895& \textbf{0.948}& 0.958& 0.629& 0.717& 0.513& 0.562\\
        FATE$^*$ & \textbf{0.911} & \textbf{0.926} & 0.918 & 0.951 & 0.484 & 0.683 & 0.517 & 0.529 \\
        \textit{RLAE} & 0.792& 0.829 & 0.843& 0.840 & 0.611& 0.657& 0.521& 0.542\\
        \textit{RoSAE} & 0.907 & \textbf{0.926} & 0.939 & \textbf{0.983} & \textbf{0.713} & \textbf{0.813} & \textbf{0.560} & \textbf{0.594}\\
 \hline
 \multicolumn{9}{c}{\textbf{Contextual anomalies ($\nu = 0.1$)}}\\
 \hline
         \multicolumn{1}{c}{\textbf{Model}} & \multicolumn{2}{c}{\textbf{20New.}} & \multicolumn{2}{c}{\textbf{Reuters}} & \multicolumn{2}{c}{\textbf{WOS}} & \multicolumn{2}{c}{\textbf{DBpedia}}  \\
         \hline
        \multicolumn{1}{l}{} & AP & AUC & AP & AUC & AP & AUC & AP & AUC \\
        \hline
        OCSVM & 0.220 & 0.681 & 0.471 & 0.755 & 0.599 & 0.889 & 0.566 & 0.882 \\
        RandNet & 0.164 & 0.579 & 0.337 & 0.662 & 0.147 & 0.544 & 0.319 & 0.694 \\
        RSRAE & 0.177 & 0.584 & 0.340 & 0.729 & 0.214 & 0.623 & 0.260 & 0.660 \\
        CVDD & 0.131& 0.613& 0.481& 0.824& 0.623& 0.916& 0.646& 0.919\\
        FATE$^*$ & 0.301 & 0.689 & 0.536 & 0.867 & 0.628 & 0.913 & 0.784 & \textbf{0.951} \\
        \textit{RLAE} & 0.201 & 0.642 & 0.420 & 0.757 & 0.334 & 0.738 & 0.368 & 0.747 \\
        \textit{RoSAE} & \textbf{0.325} & \textbf{0.718} & \textbf{0.609} & \textbf{0.880} & \textbf{0.687} & \textbf{0.921} & \textbf{0.840} & \textbf{0.951} \\
        \hline
    \end{tabular}
    \caption{Independent and contextual contamination results with contamination rate $\nu = 0.10$. Area under ROC (AUC) is the default evaluation metric for independent contamination and Average Precision (AP) is also added for contextual contamination. Embedding is Distill RoBERTA and each result is performed on test split through Algorithm~\ref{alg:tac}. $^*$ FATE uses few-shot learning and has awareness of 5 anomalies.}
    \label{tab:results_global}
\end{table*}
\begin{figure*}[t!]
	\hfill
	\begin{subfigure}[b]{.24\textwidth}
		\centering
		\includegraphics[width=\textwidth]{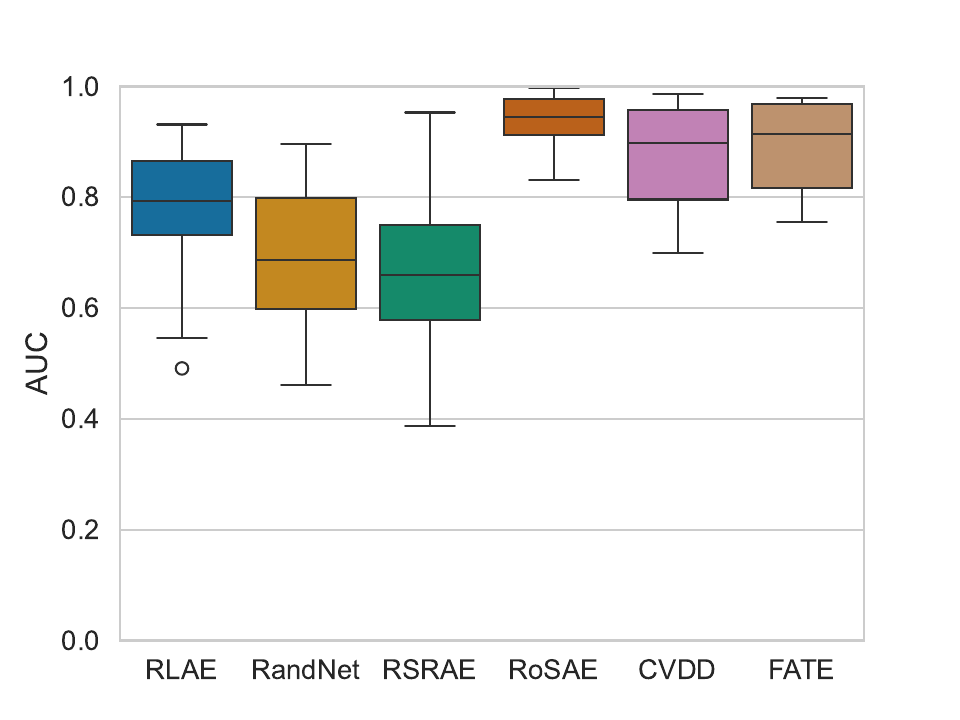}
		\caption{DBpedia 14}
		\label{fig:ensemble_db_ac}
	\end{subfigure}
	\begin{subfigure}[b]{.24\textwidth}
		\centering
		\includegraphics[width=\textwidth]{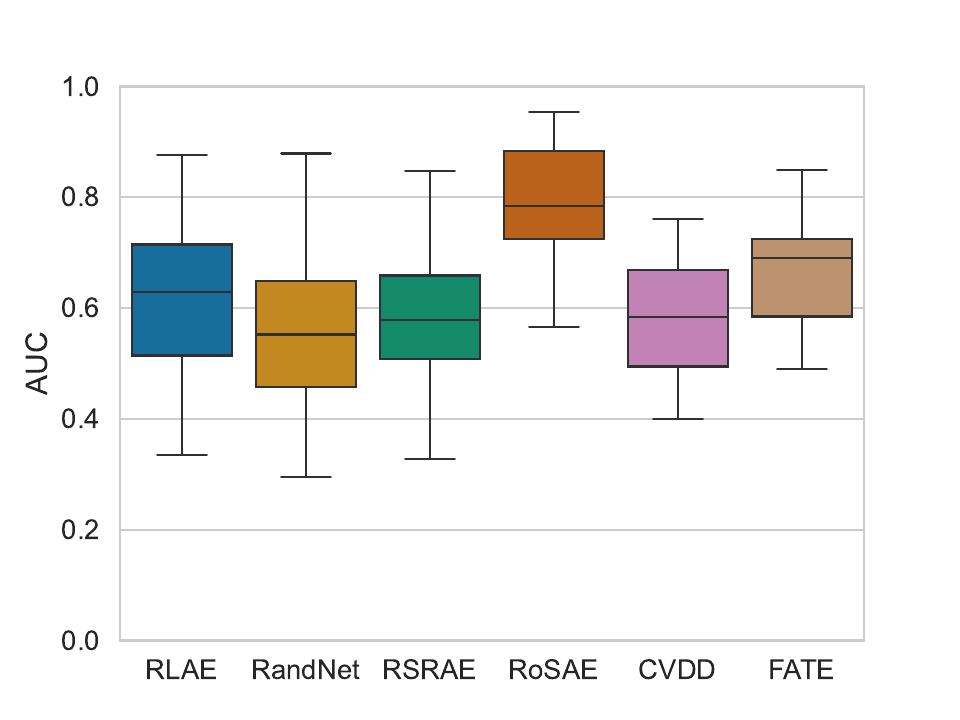}
		\caption{20 Newsgroups}
		\label{fig:ensemble_ns_ac}
	\end{subfigure}
	\hfill 
	\begin{subfigure}[b]{.24\textwidth}
		\centering
		\includegraphics[width=\textwidth]{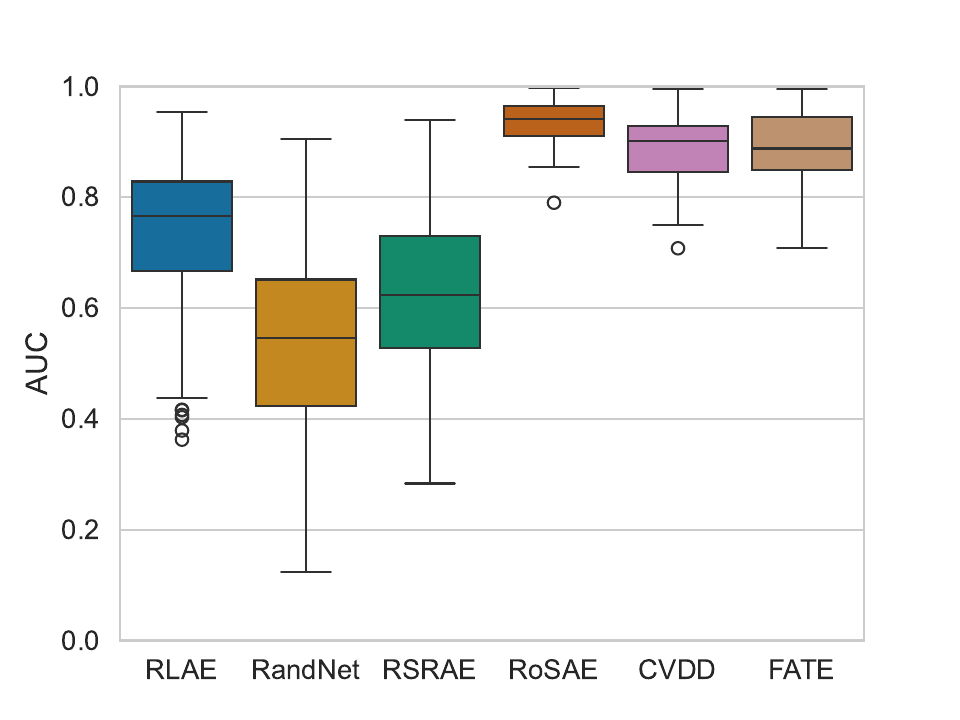}
		\caption{Web of Science}
		\label{fig:ensemble_wo_ac}
	\end{subfigure}
	\begin{subfigure}[b]{.24\textwidth}
		\centering
		\includegraphics[width=\textwidth]{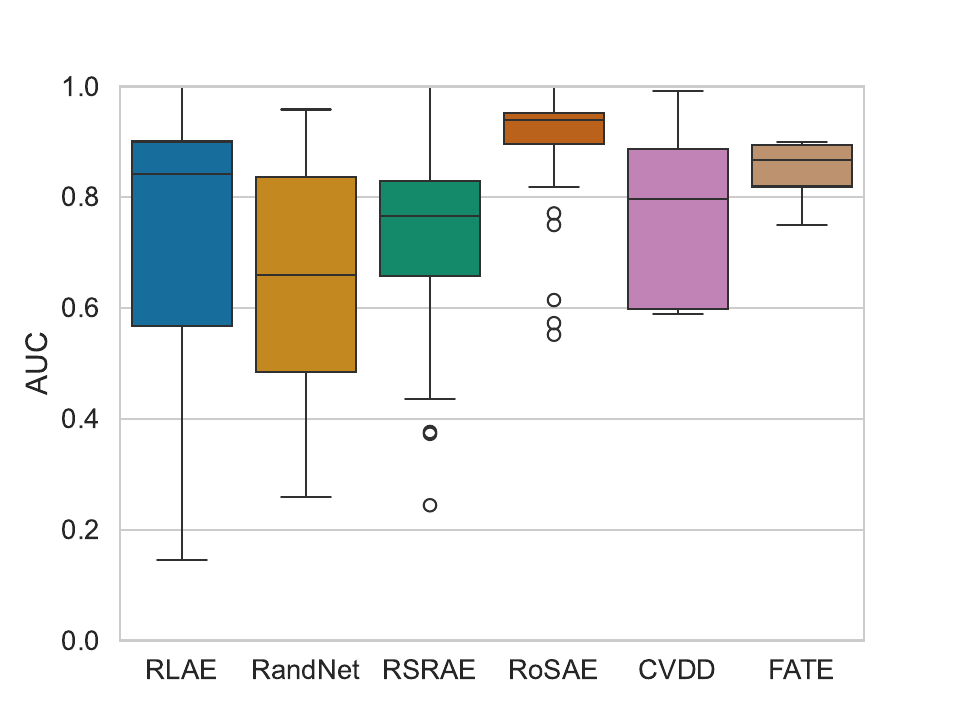}
		\caption{Reuters-21578}
		\label{fig:ensemble_rt_ac}
	\end{subfigure}
	\hfill
	\caption{Experimental study (AUC) with $\nu = 0.05$, split size of $350$ and number of base detector of $25$.}
	\label{fig:discuss_ensemble_ac}
\end{figure*}
\begin{figure*}[t!]
	\hfill
	\begin{subfigure}[b]{.24\textwidth}
		\centering
		\includegraphics[width=\textwidth]{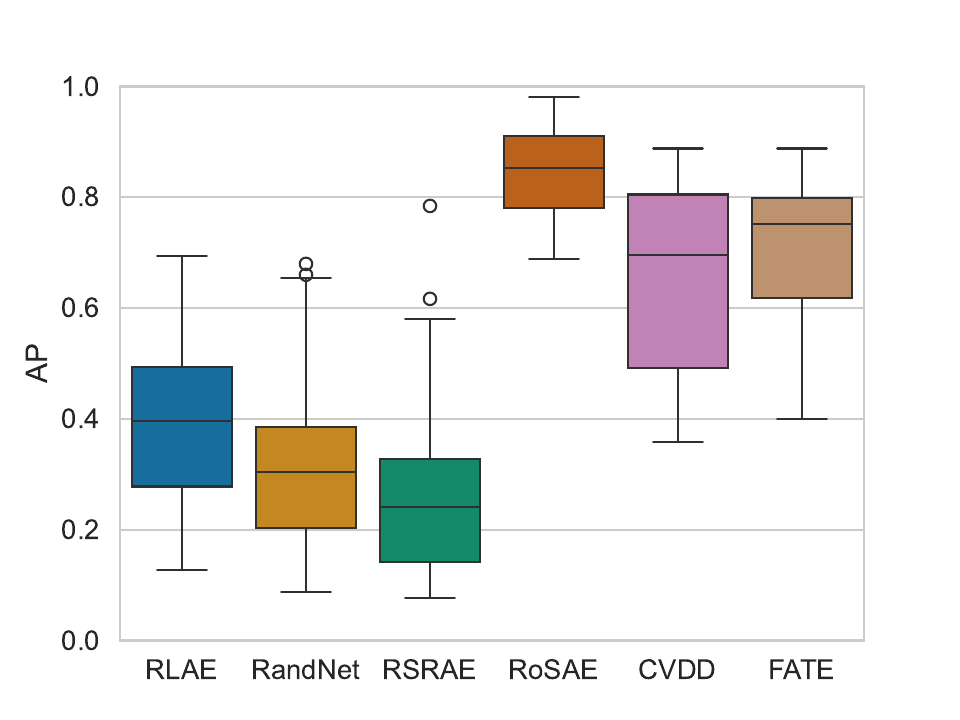}
		\caption{DBpedia 14}
		\label{fig:ensemble_db_ap}
	\end{subfigure}
	\begin{subfigure}[b]{.24\textwidth}
		\centering
		\includegraphics[width=\textwidth]{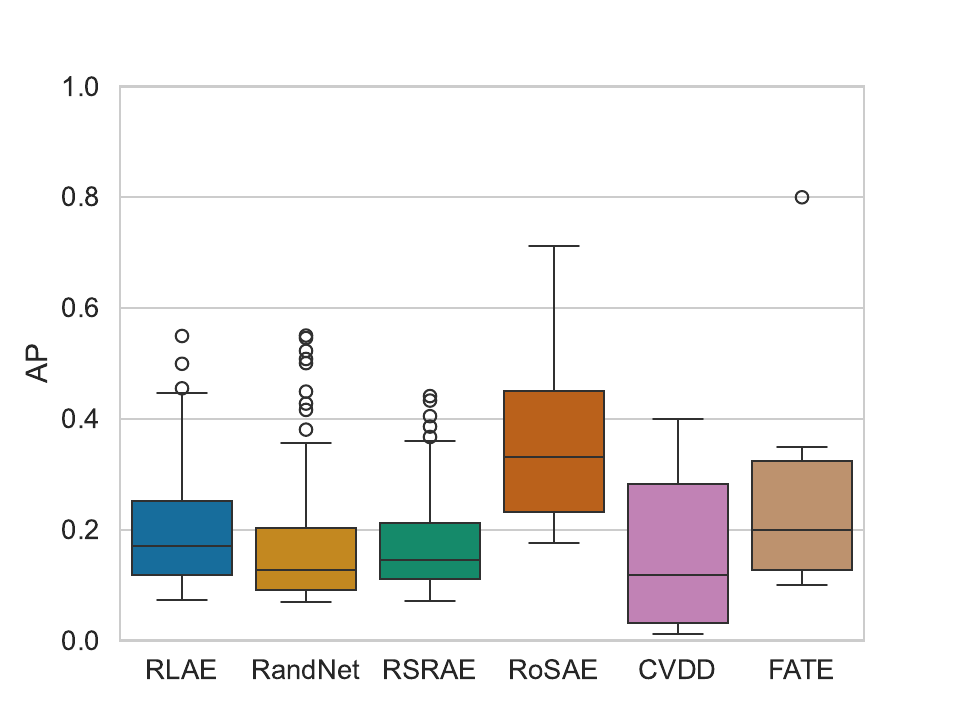}
		\caption{20 Newsgroups}
		\label{fig:ensemble_ns_ap}
	\end{subfigure}
	\hfill 
	\begin{subfigure}[b]{.24\textwidth}
		\centering
		\includegraphics[width=\textwidth]{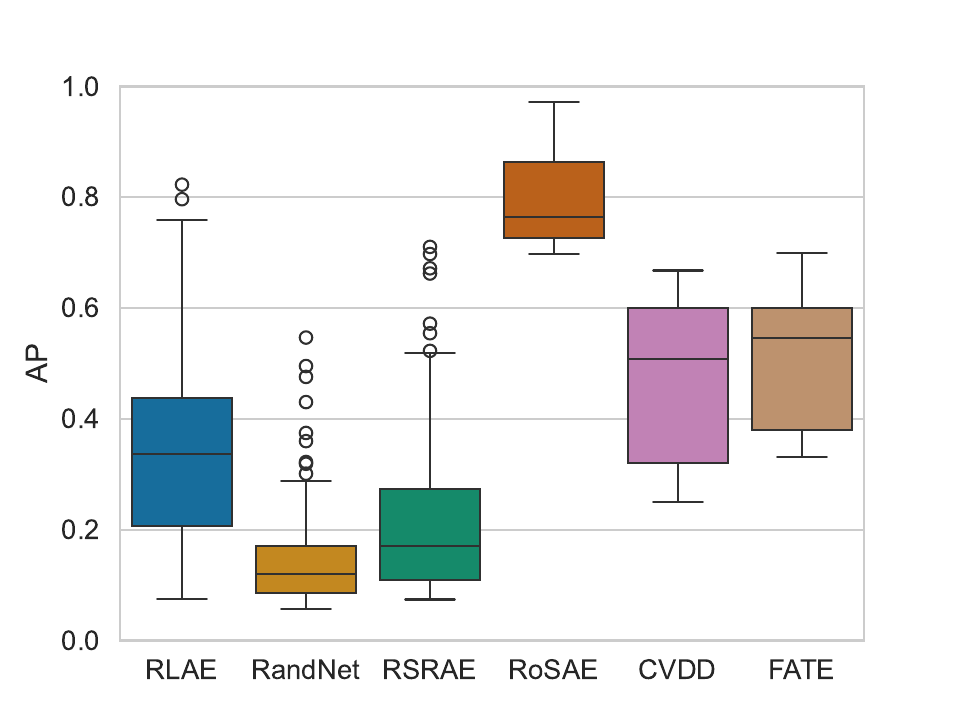}
		\caption{Web of Science}
		\label{fig:ensemble_wo_ap}
	\end{subfigure}
	\begin{subfigure}[b]{.24\textwidth}
		\centering
		\includegraphics[width=\textwidth]{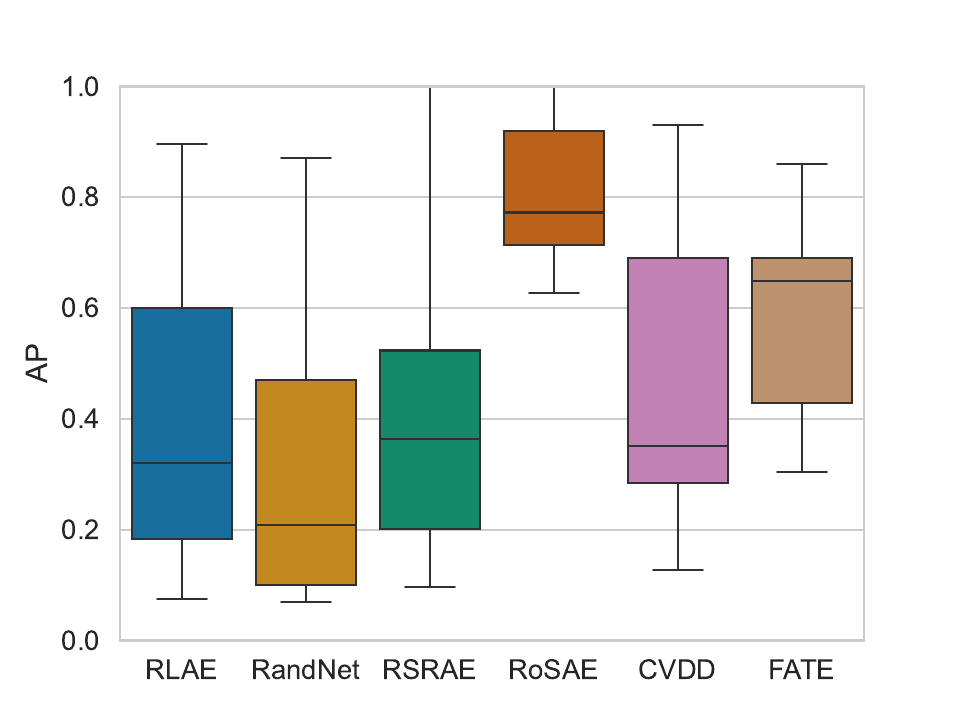}
		\caption{Reuters-21578}
		\label{fig:ensemble_rt_ap}
	\end{subfigure}
	\hfill
	\caption{Experimental study (AP) with $\nu = 0.05$, split size of $350$ and number of base detector of $25$.}
	\label{fig:discuss_ensemble_ap}
\end{figure*}
\subsection{Results} 
We propose to present our results on three principal points: independent contamination, contextual contamination and robustness of model scores.
Table~\ref{tab:results_global} displays results on TAC and independent contamination.
We can see that RoSAE succeeds to compete against recent contributions like FATE.
While the results are similar, our approach is notably standing over the others on non topic-related corpora.
Thus, our ensemble method presents success for semantic related, spam and sentiment corpora on independent contamination.

For contextual contamination we observe that our approach is outperforming others model with AUC metric and AP metric.
We can see that usage of RoSAE allows to mitigate unstable decision of the original RLAE.
We can also see significant difference of performance with 20 Newsgroups corpus and Reuters-21578.
For contextual contamination, our approach outperforms all models.
On AP metric, RoSAE presents 9\% higher results than top competitor, on average.

\begin{figure*}[t!]
	\hfill
	\begin{subfigure}[b]{.49\textwidth}
		\centering
		\includegraphics[width=\textwidth]{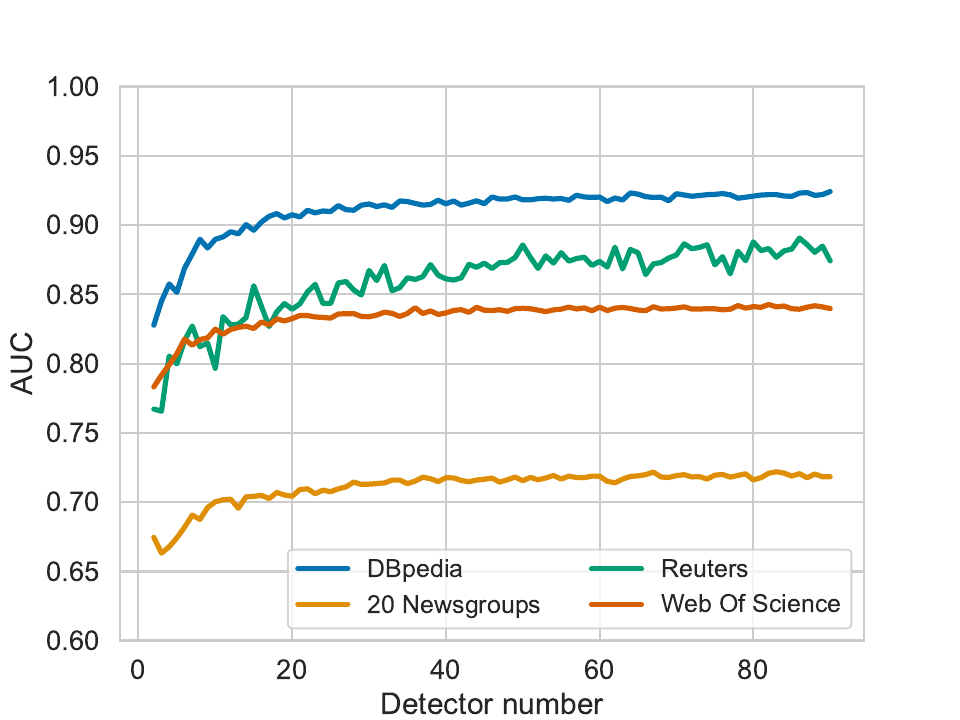}
		\caption{Number of base detector}
		\label{fig:detector}
	\end{subfigure}
	\begin{subfigure}[b]{.49\textwidth}
		\centering
		\includegraphics[width=\textwidth]{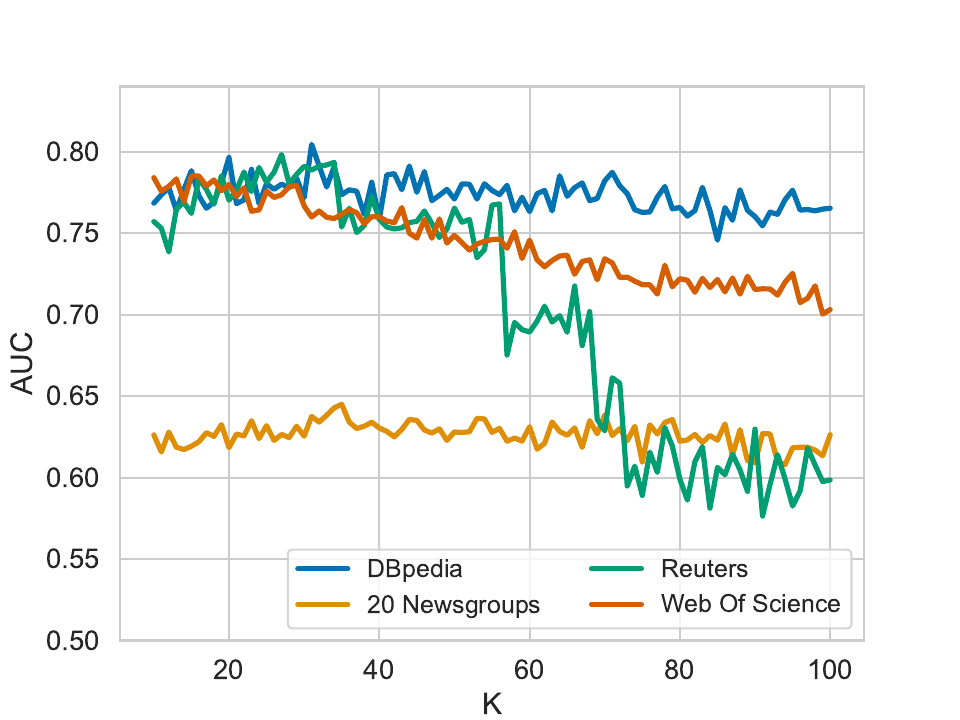}
		\caption{Study of hyperparameter K}
		\label{fig:parameterk}
	\end{subfigure}
	\hfill
	\caption{Results of RoSAE with contextual contamination on size of ensemble and on different values for hyperparameter K.}
	\label{fig:discuss_detector}
\end{figure*}

\begin{figure*}[t!]
	\hfill
	\begin{subfigure}[b]{.32\textwidth}
		\centering
		\includegraphics[width=\textwidth]{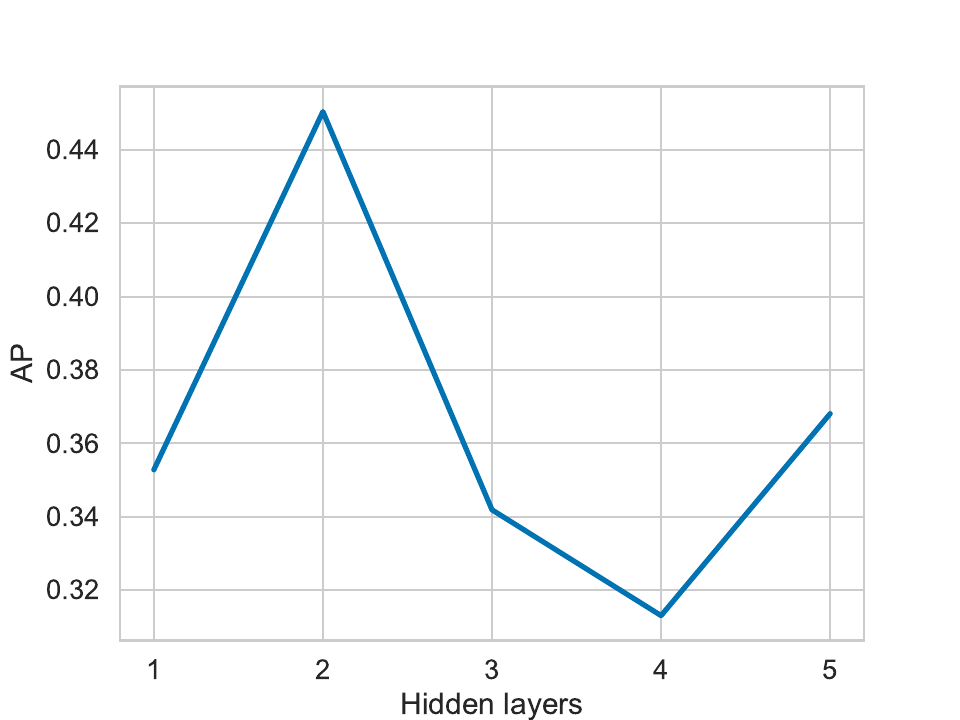}
		\caption{Hidden layers}
		\label{fig:hidden}
	\end{subfigure}
	\hfill
	\begin{subfigure}[b]{.32\textwidth}
		\centering
		\includegraphics[width=\textwidth]{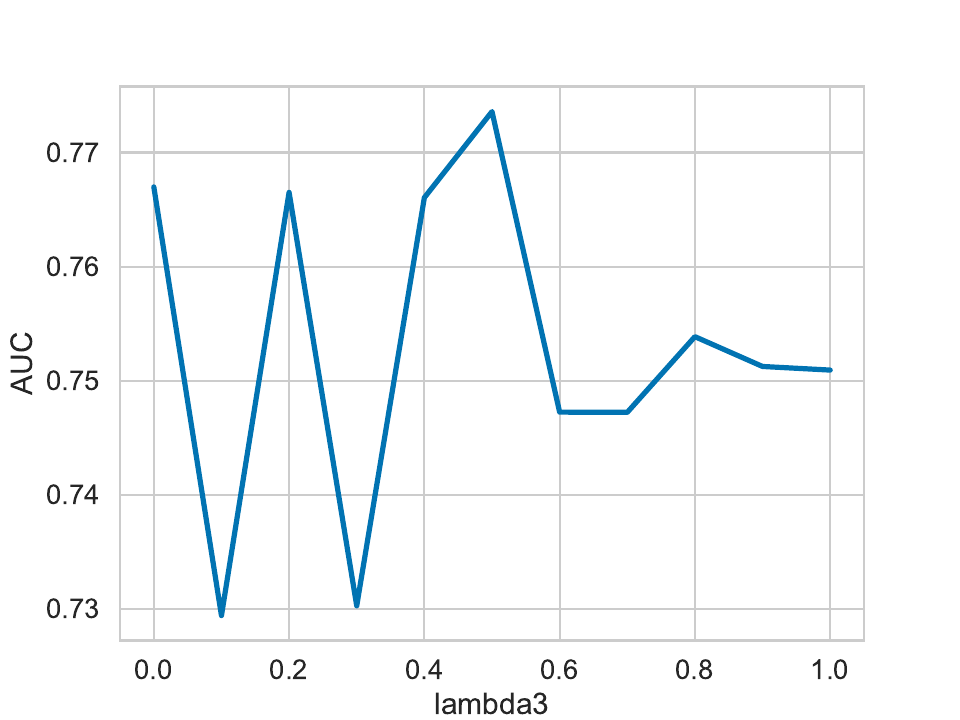}
		\caption{Lambda}
		\label{fig:lambda}
	\end{subfigure}
	\hfill 
	\begin{subfigure}[b]{.32\textwidth}
		\centering
		\includegraphics[width=\textwidth]{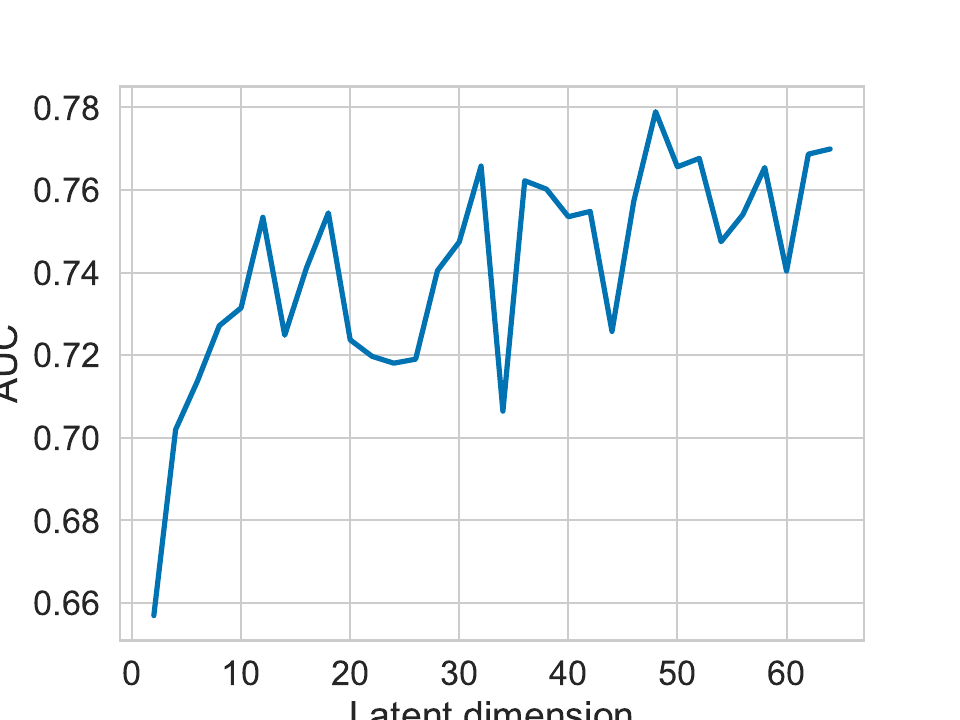}
		\caption{Latent space}
		\label{fig:latent}
	\end{subfigure}
	\hfill
	\caption{Study of RLAE hyperparameter: hidden layers number, lambda3 regulating local recovery term and latent space dimension.}
	\label{fig:discuss_ablation}
\end{figure*}

While our performances are competitive, the principal purpose of tackling anomaly detection with ensemble methods is to mitigate the bias-variance trade-off.
For benchmarking the robustness of each method, we use a limited amount of documents in training and a low contamination rate ($\nu = 0.05$), presenting a difficult scenario to handle.
The Figure~\ref{fig:discuss_ensemble_ac} and the Figure~\ref{fig:discuss_ensemble_ap} displays outperforming results from our approach.
We can see that the variance of our model is noticeable as the box variance are always similar or smaller than its competitors.
Also, the min and max possible scores are close from the median scores, concluding to see that our approach is more efficient and more robust.

Reference results are considerably different from ours.
The principal reason is due the the contamination process for performing experimental study.
We can observe that the literature contaminate both independent anomalies and contextual anomalies without distinctive analysis.
Another observation lies in the competitive results from FATE thanks to its few-shot learning mechanism.

\section{Ablation study}
In this section we propose an observation of various properties of our autoencoder RLAE and its impact inside RoSAE (ensemble).

\paragraph{Base detector number}
The Figure~\ref{fig:detector} displays the performance impact against the base detector number in RoSAE.
We can see that starting from 30, there is not substantial gains to increase the number of base detector in the ensemble.
While it can slowly increase AUC, it considerably increase the computation time.
20 is then a concrete spot.

\paragraph{Number of local neighbours}
In Figure~\ref{fig:parameterk} we display the impact of the hyperparameter $k$ in RLAE.
As stated in the Reuters-21578 results, we can see that this hyperparameter needs to be wisely picked.
While a low value seems to be preferable, we can see that our autoencoder is still robust against this hyperparameter for other dataset.

\paragraph{Parameters}
In Figure~\ref{fig:discuss_ablation} is presented sensibility to three hyperparemeter that impact the learning step: number of hidden layer, the $\lambda_3$ parameter and the size of latent space in the RoSAE model.
While variance is tackled in ensemble methods, hidden layers are not a highly sensible parameter considering the computation/performance tradeoff.
In Section~\ref{sec:expe} the hyperparameter $\lambda_3 < \lambda_1$ is set to $0.05$.
The curve is the AUC performance against $\lambda_3$ for $\lambda_1 = 0.1$.
Finally, the learned embedding inside RLAE should be preferably lower than the output dimension of the autoencoder.

\section{Conclusion}
\label{sec:conclusion}
In this work we have introduced RoSAE, an ensemble approach with robust autoencoders, optimized through LNE for tackling contextual anomaly in text.
Conducted experiments demonstrate that our approach outperform all baseline methods with less variance on contextual contamination.
We can also note that results displays that state-of-the-art approaches are performing independent contamination, and present less robust results than RoSAE.

\section{Limitations and Ethical Considerations}
We admit one major limitation from this work.
As we have presented a large number of dataset, we lack of dedicated corpora.
While our work is an attempt to reproduce real-world scenario, we only rely to synthetic contamination.
On the other hand, anomaly detection is useful for users and experts, but can lead to unfortunate usages in social media.




\bibliography{anthology,main}

\end{document}